\def\year{2018}\relax
\documentclass[letterpaper]{article} 
\usepackage{aaai18}  
\usepackage{times}  
\usepackage{helvet}  
\usepackage{courier}  
\usepackage{url}  
\usepackage{graphicx}  
\usepackage[utf8]{inputenc} 
\usepackage[T1]{fontenc}    
\usepackage{booktabs}       
\usepackage{amsfonts}       
\usepackage{nicefrac}       
\usepackage{microtype}      

\usepackage{graphicx}
\graphicspath{{./figures/}}
\usepackage{subcaption}
\usepackage[ruled]{algorithm2e}
\usepackage{caption}
\usepackage{color}
\usepackage{ctable}
\usepackage{siunitx}
\usepackage{amsmath}
\usepackage{enumitem}
\usepackage{multirow}

\title{Quantized Memory-Augmented Neural Networks}
\author{Seongsik Park\textsuperscript{1}, Seijoon Kim\textsuperscript{1}, Seil Lee\textsuperscript{1}, Ho Bae\textsuperscript{2}, and Sungroh Yoon\textsuperscript{1,2\textdagger}\\
\textsuperscript{1}Department of Electrical and Computer Engineering, Seoul National University, Seoul 08826, Korea\\
\textsuperscript{2}Interdisciplinary Program in Bioinformatics, Seoul National University, Seoul 08826, Korea\\ 
\textsuperscript{\textdagger}sryoon@snu.ac.kr
}

\begin{document}

\maketitle

\begin{abstract}
Memory-augmented neural networks (MANNs) refer to a class of neural network models equipped with external memory (such as neural Turing machines and memory networks).
These neural networks outperform conventional recurrent neural networks (RNNs) in terms of learning long-term dependency, allowing them to solve intriguing AI tasks that would otherwise be hard to address.
This paper concerns the problem of quantizing MANNs.
Quantization is known to be effective when we deploy deep models on embedded systems with limited resources.
Furthermore, quantization can substantially reduce the energy consumption of the inference procedure.
These benefits justify recent developments of quantized multilayer perceptrons, convolutional networks, and RNNs.
However, no prior work has reported the successful quantization of MANNs.
The in-depth analysis presented here reveals various challenges that do not appear in the quantization of the other networks.
Without addressing them properly, quantized MANNs would normally suffer from excessive quantization error which leads to degraded performance.
In this paper, we identify memory addressing (specifically, content-based addressing) as the main reason for the performance degradation and propose a robust quantization method for MANNs to address the challenge.
In our experiments, we achieved a computation-energy gain of 22$\times$ with 8-bit fixed-point and binary quantization compared to the floating-point implementation.
Measured on the bAbI dataset, the resulting model, named the quantized MANN (Q-MANN), improved the error rate by 46\% and 30\% with 8-bit fixed-point and binary quantization, respectively, compared to the MANN quantized using conventional techniques.






\end{abstract}

\section{Introduction}
In recent years, Memory-Augmented Neural Networks (MANNs), which couple the external memory and the neural network, have been proposed for improving the learning capability of the neural network.
MANNs, including neural Turing machines~\cite{graves2014neural}, memory networks~\cite{weston2014memory}, and differentiable neural computers~\cite{graves2016hybrid}, can handle complex problems such as long-sequence tasks and Question and Answer (QnA).

Although various types of neural networks have shown promising results in many applications, they demand tremendous amounts of computational energy~\cite{silver2016mastering}.
This has been the biggest obstacle for employing deep learning on embedded systems with limited hardware resources.

Previous work overcame this challenge by introducing the concept of quantization (e.g., fixed-point and binary) to deep learning~\cite{courbariaux2014training,lin2015fixed,gupta2015deep,hubara2016quantized,courbariaux2015binaryconnect,courbariaux2016binarized,rastegari2016xnor,zhou2016dorefa,tang2017train}.

Quantization can avoid floating-point operations which consume considerable computing energy~\cite{horowitz20141}.
As shown in Table~\ref{tab:arithmetic_operation_energy}, an 8-bit fixed-point addition can obtain a computation-energy gain of 123.3$\times$ over a 32-bit floating-point multiplication (Supplementary Table~\ref{tab:sup_arithmetic_operation_energy}).

Typically, low-power and real-time processing are recommended features in a limited-resource environment. NVIDIA has recently introduced TensorRT~\cite{nvidia2017}, a deep learning inference optimizer for low-power and real-time processing with 8-bit or 16-bit representation. For the reasons mentioned above, quantization has become a critical process to employ deep learning in a limited-resource environment. 

\ctable[
pos = t,
caption = {\fontsize{9.0pt}{9.0pt} \selectfont Computation-energy consumption from~\cite{horowitz20141}},
width=\linewidth,
label = {tab:arithmetic_operation_energy},
doinside = {\footnotesize \def\arraystretch{.7}}
]{@{\extracolsep{\fill}}lcccc}{
\tnote[a]{ compared with 32-bit floating-point mult }
}{ \FL
    \multirow{2}{*}{Type} & Arithmetic & \multirow{2}{*}{Bit} & Energy & \multirow{2}{*}{Gain\tmark[a]} \NN
    & operation & & (pJ) &  \ML
    Fixed & add & 8 & 0.03 & 123.3$\times$ \NN
 
	point & mult & 8 & 0.2 & 18.5$\times$ \ML
	Floating & add & 32 & 0.9 & 4.1$\times$ \NN
	point & mult & 32 & 3.7 & 1$\times$ \LL
}
In this paper, we applied fixed-point and binary quantization to a conventional MANN to enable it to be employed in a limited-resource environment.
We evaluated its feasibility by training and inference with fixed-point quantized parameters and activations, and then further extended this to binary representation.
According to our experimental results, the application of 8-bit fixed-point quantization increased the test error rate by more than 160\%.
This was a significant amount of degradation compared to that of the initial study of fixed-point quantization on CNN~\cite{lin2016overcoming}.

This prompted us to overcome the limitations of training and inference with quantized parameters and activations in a MANN.
We started by theoretically analyzing the effect of the quantization error on the MANN using content-based addressing with cosine similarity.
We verified our analysis and revealed that a conventional MANN was susceptible to quantization error through various experiments.

Based on our detailed analysis and experiments, we proposed quantized MANN (Q-MANN) that employs content-based addressing with a Hamming similarity and several techniques to enhance its performance. 
Our result showed that Q-MANN had a lower test error rate than conventional MANN when both 8-bit fixed-point and binary quantization were applied. The contributions of this paper can be summarized as follows:

\begin{itemize}[topsep=0pt,itemsep=0ex,partopsep=1ex,parsep=1ex,leftmargin=*]
    \item We first attempted to train a conventional MANN with fixed-point and binary quantized parameters and activations, which could be the basis for subsequent research.
    \item We theoretically analyzed the reasons for the poor training result when applying quantization to the MANN and verified our analysis through various experiments. 
    \item We proposed Q-MANN that could improve both the robustness in the quantization error and the learning performance when a small bit width of quantization was applied.
\end{itemize}

\section{Related Work}
\subsection{Memory-Augmented Neural Networks}
A MANN consists of two main components: a memory controller and external memory.
It can learn how to read from and write to the memory through data.
Thus, memory addressing plays a key role in training and inference.

There could be several types of addressing method depending on the interpretation of MANN.
However, in this paper, we referred to MANN as neural networks represented by neural Turing machines~\cite{graves2014neural}, differentiable neural computers~\cite{graves2016hybrid} and memory networks~\cite{weston2014memory}. 
Thus, we focused on content-based addressing which is general addressing method of those neural networks.

Content-based addressing $C$ is defined as
\begin{equation}
\label{eq:content_based_addr}
C(M,k)[i] = \frac{\exp\{S(M_i,k)\}}{\sum_{j=1}^{L}{\exp\{S(M_j,k)\}}}\textrm{,}
\end{equation}
where $S$ is the similarity between the memory element $M_i$ and the key vector $k$, and $L$ is the number of memory elements.
In a conventional MANN, a cosine similarity is used as a similarity measure $S$ between the memory element $M_i$ and the key vector $k$, and a softmax function is used as a normalization function.
Since cosine similarity and softmax are differentiable, MANN is end-to-end trainable using SGD and gradient back-propagation.

However, this approach is not suitable in a limited-resource environment because the cosine similarity that features multiplication, division, and square root computation require tremendous amounts of hardware resources and computational energy.
In this regard, using a dot product instead of cosine similarity as in~\cite{sukhbaatar2015end} is much more efficient in a limited-resource environment.

\subsection{Fixed-point and binary quantization}
Fixed-point quantization is a popular choice because it hardly has any quantization overhead and is seamlessly compatible with conventional hardware. Given a floating point $u$, the fixed point $\hat{u}$ can be expressed as
\begin{equation}
\label{eq:fixed_point_quant}
    \hat{u}= S_{\hat{u}}2^{-FRAC} \sum _{k=0}^{n-2}{2^k \hat{u}_{k}}\textrm{,}
\end{equation}
where $n$ is the bit width, $S_{\hat{u}}$ is the sign, and $FRAC$ is the fraction bit of fixed point $\hat{u}$. The fixed point is composed of a sign, integer, and fraction and each bit width can be denoted as 1, $IWL$, and $FRAC$.
Q-format (Q\scriptsize$IWL$\normalsize.\scriptsize$FRAC$\normalsize) was used to represent a fixed point in this paper.

The quantization error $\epsilon_{\hat{u}}$ is defined as depending on the occurrence of fixed-point overflow,
\begin{equation}
\label{eq:fixed_point_quant_error}
    |\epsilon_{\hat{u}}| < 
    \begin{cases}
        2^{-FRAC}       & \textrm{if}~|u|<2^{IWL} \\
        |2^{IWL}-|u||   & \textrm{if}~|u|\geq2^{IWL} \ \textrm{.}
    \end{cases}
\end{equation}

More details of the fixed point are provided in Supplementary.
As shown in Equation~\ref{eq:fixed_point_quant_error}, when a fixed-point overflow occurs, the quantization error becomes much larger and this can be a major cause of the degradation of the learning performance of a MANN.

As shown in Table~\ref{tab:arithmetic_operation_energy}, an energy gain of 18.5$\times$ could be accomplished through fixed-point quantization by converting 32-bit floating-point arithmetic operations to 8-bit fixed-point. 
Moreover, we could achieve a 123.3$\times$ energy gain by replacing 32-bit floating-point multiplication with 8-bit fixed-point addition through a binary quantization.
These two approaches allowed us to deploy a MANN in a limited-resource environment.

\subsection{Fixed-point and binary quantization on deep learning}

To date, most of the research on fixed-point and binary quantization of deep learning has focused on MLP, CNN, and RNN~\cite{hubara2016quantized}.
Especially, quantization on CNN has been extended beyond fixed-point quantization to binary quantization \cite{courbariaux2015binaryconnect,courbariaux2016binarized,rastegari2016xnor,tang2017train}.

Initial studies on fixed-point quantization of CNN include \cite{gupta2015deep} and \cite{lin2016overcoming}.
By applying 16-bit fixed-point quantization and using stochastic rounding, \cite{gupta2015deep} could obtain a similar result to that of a floating-point on the CIFAR-10 dataset.
\cite{lin2016overcoming} confirmed the feasibility of fixed-point quantization on CNN using fewer bits by applying 4-/8-bit fixed-point quantization to CNN.

Based on the initial research, binary quantization on CNN has been studied beyond fixed-point quantization.
In BinaryConnect~\cite{courbariaux2015binaryconnect}, binary quantization was applied to the parameters in CNN.
It showed a comparable result to floating point on the MNIST, CIFAR-10, and SVHN datasets.
Through binary quantization, they were able to reduce the computation-energy consumption by replacing multiplication with addition.

BinaryNet~\cite{courbariaux2016binarized} and XNOR-Net~\cite{rastegari2016xnor} applied binary quantization of parameters and activation to CNN.
The binarization method was the same as that of BinaryConnect, and the estimated gradient was used for gradient back-propagation.
Multiplication was replaced with XNOR and addition, which allowed training and inference with less computation-energy consumption than BinaryConnect.

BinaryNet achieved similar results to BinaryConnect on the MNIST and CIFAR-10 datasets. XNOR-Net used a scale factor and adjusted the position of the layers to compensate for the loss of information due to binarization. As a result, binary quantization could be applied to CNN on the ImageNet dataset with an increase in the error rate of approximately 50\% compared to the use of floating point. DoReFa-Net~\cite{zhou2016dorefa} with a quantized gradient showed a lower error rate than XNOR-Net on the ImageNet dataset when the parameters, activation, and gradient were 1-, 2-, and 4-bit, respectively.

The most recently published research~\cite{tang2017train} could improve the learning performance of binarized CNN by using a low learning rate, PReLU instead of a scale factor, and newly proposed regularizer for binarization.
They obtained an increase in the error rate of about 20\% compared to the using floating point on the ImageNet dataset, which is the lowest error rate thus far.

\cite{ott2016recurrent} mainly focused on binary and ternary quantization on RNN. They applied quantization to parameters of vanilla RNN, LSTM, and GRU.
In their experiments, they obtained a comparable result to those using floating point, but only when the ternarization was applied.


\section{In-Depth Analysis of the Quantization Problem on a MANN}\label{probme-state}
\subsection{Fixed-point and binary quantization on a MANN}
\begin{figure*}[tbp]
    \centering
    \includegraphics[width=1.0\linewidth]{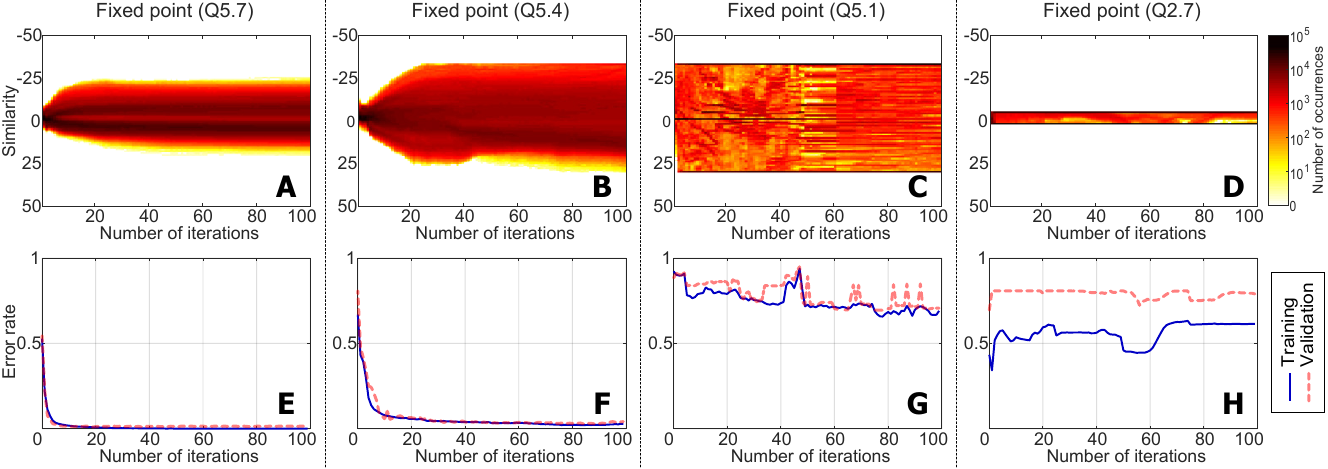}
	\caption{ The results for MANN on the bAbI dataset (task8), (A)-(D): the distribution of similarity measure, (E)-(H): the error rate of training and validation.
	The results of measuring the distribution of the similarity by setting $IWL$ to 5 and reducing $FRAC$ to 7, 4, and 1 (A, B, and C, respectively) showed an increase in the width of distribution as $FRAC$ decreased. In addition, the width of distribution became wider as the training progressed.
	The increase of the width caused the fixed-point overflow because a fixed point could represent a value in a limited range ($<|2^{IWL}|$).
	If the overflow occurs frequently as (C) and (D), the error rate of training and validation increased considerably as (G) and (H).
	Even if the same bit-width was used for the fixed-point quantization as (B) and (D), the learning performance varied greatly, as (F) and (H), according to the width of $IWL$ and $FRAC$.
	}
	\label{fig:MANN_similarity_lr_curve_one}
\end{figure*}

To the best of our knowledge, this work is the first attempt to apply fixed-point and binary quantization to a MANN.
Our method involved applying the approaches that are conventionally used to achieve fixed-point and binary quantization in CNNs to a MANN.
Our approach was similar to that in~\cite{gupta2015deep,courbariaux2015binaryconnect} and we analyzed the feasibility of applying fixed-point and binary quantization to a MANN.
As the early stage research of quantization on CNN, we excluded the last output layer and gradient from quantization and focused on the activation and parameters instead.

We followed the conventional approach to fixed-point quantization by quantizing the parameters, activation, and memory elements to 8-bit fixed point.
For binary quantization, we set the activation function as binary, and the parameters and memory elements as 8-bit fixed point. We binarized the activation instead of the parameters since the input format (e.g., Bag-of-Words) of MANN can be easily binarized and the parameter can directly affect the memory element significantly. In those respects, our binarization method is different from \cite{courbariaux2015binaryconnect}, but it is similar in that it can reduce computation-energy consumption by replacing multiplication with addition.

Our experimental results indicated that a conventional MANN increased the error rate by over 160\% when 8-bit fixed-point quantization was applied.
Compared with the result obtained during the initial stage research of quantization on CNN~\cite{lin2016overcoming}, which had an increased error rate of 40\%, the learning performance achieved by applying quantization to the conventional MANN was significantly degraded.
In this section, we thoroughly investigate why the quantization affected serious performance degradation in the conventional MANN. Based on the analysis, our proposed approach to solving the stated problem is presented in the later section.

\subsection{Analysis of the effect of the quantization error on content-based addressing}

One of the major causes of performance degradation is content-based addressing which is an external memory addressing technique. Content-based addressing typically consists of a cosine similarity (i.e., dot product) and a softmax function (Equation~\ref{eq:content_based_addr}).

The distribution of vector similarity used in content-based addressing of a conventional MANN is depicted in Figure~\ref{fig:MANN_similarity_lr_curve_one}. 
As the training progressed, the similarity of the related vectors became larger and that of the less relevant vectors became smaller. Thus, the more training progressed, the wider the distribution of similarity became. This training tendency of the vector similarity was a major obstacle to applying fixed-point quantization to the conventional MANN.

We analyzed the effect of the quantization error on the distribution of the similarity by calculating the similarity $\hat{Z}$,
\begin{equation}
\label{eq:quant_effect_dot_product}
    \hat{Z} \approx Z + \sum {(u_i \epsilon_{\hat{v}_i} + v_i \epsilon_{\hat{u}_i})}\textrm{,}
\end{equation}
where $Z$ is the similarity of floating-point vector, $\hat{u}_i~\text{and}~\hat{v}_i$ are elements of the quantized input vectors  (Supplementary Equation~\ref{eq:sup_quant_effect_dot_product}).
As shown in Equation~\ref{eq:quant_effect_dot_product}, the influence of the quantization error of the input vectors $\epsilon_{\hat{u}_i}~\text{and}~\epsilon_{\hat{v}_i}$ on the error of similarity measure $\epsilon_{\hat{Z}}$ was proportional to the sum of the products of the input vectors and the quantization error $\sum {(u_i \epsilon_{\hat{v}_i} + v_i \epsilon_{\hat{u}_i})}$. As mentioned earlier, since the distribution of similarity became wider as the training progressed, the quantization error of the input vectors $\epsilon_{\hat{u}_i}~\text{and}~\epsilon_{\hat{v}_i}$ caused the distribution of similarity to become much wider. 

In order to examine the effect of the fixed-point quantization error on the distribution of similarity, we set $IWL$ to 5 and measured the distribution of similarity while decreasing $FRAC$ to 7, 4, and 1, respectively (Figures~\ref{fig:MANN_similarity_lr_curve_one}A, ~\ref{fig:MANN_similarity_lr_curve_one}B, and~\ref{fig:MANN_similarity_lr_curve_one}C). Since the fixed-point quantization error is inversely proportional to the $FRAC$ (Equation~\ref{eq:fixed_point_quant_error}), the distribution became wider as the $FRAC$ decreased. The wide distribution of similarity due to the quantization error incurred fixed-point overflow (Figure~\ref{fig:MANN_similarity_lr_curve_one}C), which made conventional MANN vulnerable to a quantization error (Figure~\ref{fig:MANN_similarity_lr_curve_one}G).

To investigate the effect of the error of similarity measure on the learning performance of a conventional MANN, we analyzed the effect of the input quantization error $\epsilon_{\hat{z}_i}$ on the output of a softmax $\hat{y}_i$ which is a typical normalization function in content-based addressing:
\begin{equation}
\label{eq:quant_effect_softmax}
    \hat{y}_i \leq \exp(\epsilon_{max})y_i\textrm{,}
\end{equation}
where $y_i$ is the floating-point output of softmax and $\epsilon_{max}$ is the maximum of the input quantization error (Supplementary Equation~\ref{eq:sup_quant_effect_softmax}).
Assume that index \textit{i} is referred to as the index of $\epsilon_{max}$ among the quantization error $\epsilon_{\hat{z}_i}$ of quantized input $\hat{z}_i$ in softmax.
Then, the output of softmax $\hat{y}_i$ would be proportional to $\exp(\epsilon_{max})$.
Therefore, the output error $\epsilon_{\hat{y}_i}$ would also be proportional to $\exp(\epsilon_{max})$.

This finding indicated that when softmax is used as a normalization function it is exponentially influenced by the input quantization error.
In content-based addressing, the vector similarity $Z$ was used as input of the softmax normalization function.
Thus, the error of vector similarity $\epsilon_{\hat{Z}}$ exponentially affected the output error of softmax $\epsilon_{\hat{y}_i}$.
Consequently, $\epsilon_{\hat{Z}}$ significantly degraded the learning performance of conventional MANN.

Depending on whether fixed-point overflow occurs or not, the fixed-point quantization error varies greatly according to Equation~\ref {eq:fixed_point_quant_error}.
If there were endurable overflows (Figure~\ref{fig:MANN_similarity_lr_curve_one}B), the error rate of conventional MANN was higher than that of the floating-point, but the learning tendency was not significantly different (Figures~\ref{fig:MANN_similarity_lr_curve_one}F and~\ref{fig:total_similarity_lr_curve}E).
However, the occurrence of many overflows and an increase in the error of the similarity measure (Figure~\ref{fig:MANN_similarity_lr_curve_one}C), caused the learning performance to decrease drastically (Figure~\ref{fig:MANN_similarity_lr_curve_one}G).

A comparison of our results (Figures~\ref{fig:MANN_similarity_lr_curve_one}F and~\ref{fig:MANN_similarity_lr_curve_one}H) with fixed-point quantization using the same bit width but different configurations (Q5.4 and Q2.7), enabled us to observe the degradation of learning performance due to fixed-point overflow.
Although the same bit width was used, there were many more overflows in the case of Q2.7 (Figure~\ref{fig:MANN_similarity_lr_curve_one}D), which degraded the learning performance significantly (Figure~\ref{fig:MANN_similarity_lr_curve_one}H).

Hence, fixed-point overflow had a great influence on the learning performance of MANN.
Overflow of vector similarity in fixed-point quantization where limited bit width is available can be prevented by increasing $IWL$ but this leads to the use of a smaller $FRAC$.
As mentioned earlier, the increased quantization error due to the small $FRAC$ caused fixed-point overflow of the similarity measure as training progressed (Figure~\ref{fig:MANN_similarity_lr_curve_one}C). Thus, cosine similarity (i.e., dot product) is not suitable for fixed-point quantization, which uses a small number of bits with a limited numerical representation range.

\section{Quantized MANN (Q-MANN)}\label{q-mann}
\subsection{Architecture}
\begin{figure*}[t]
    \centering
    \includegraphics[width=1.0\linewidth]{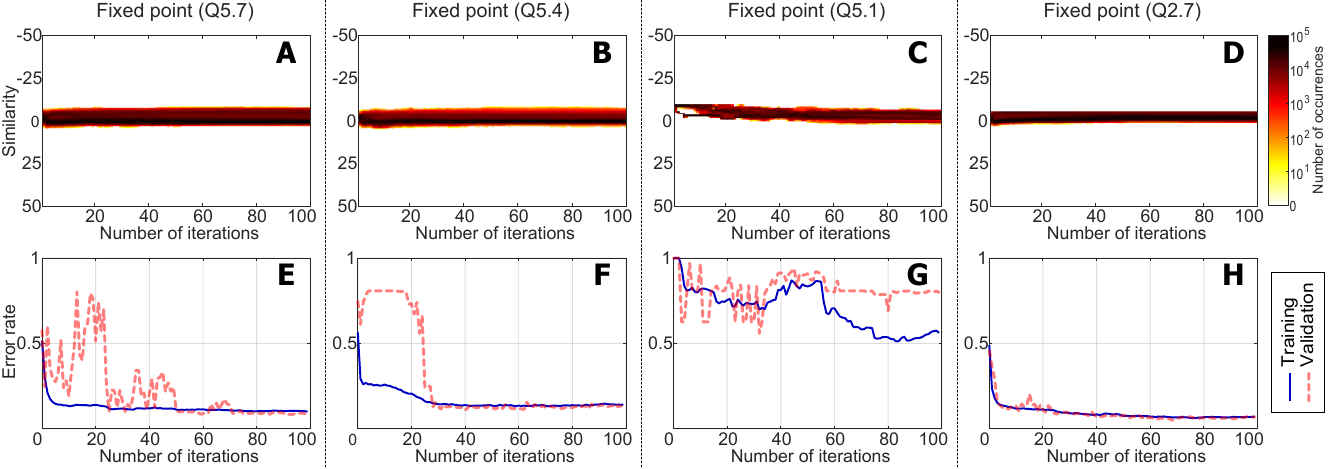}
	\caption{
	The results for Q-MANN on the bAbI dataset (task8), (A)-(D): the distribution of similarity measure, (E)-(H): the error rate of training and validation.
	The results of measuring the distribution of similarity by setting $IWL$ to 5 and reducing $FRAC$ to 7, 4, and 1 (A, B, and C, respectively), indicated that there was little difference among the width of the distributions.
	Even though there was no significant difference in the distribution of the width, the similarity measure became more accurate, and the learning performance was improved as $FRAC$ increases with an $IWL$ (E, F, and G).
	In the experiments having the same $FRAC$, the learning performance was improved when using the distribution-optimized $IWL$ as (H) ($IWL=2$), even though it was smaller than the other (E) ($IWL=5$).
	In the case of using the same bit-width, the results showed lower the error rate of training and validation with the distribution-optimized $IWL$ and $FRAC$ (Q2.7) as (H) compared to (F).
	}
	\label{fig:QMANN_similarity_lr_curve}
\end{figure*}

\begin{figure*}[t]
    \center
    \includegraphics[width=1.0\linewidth]{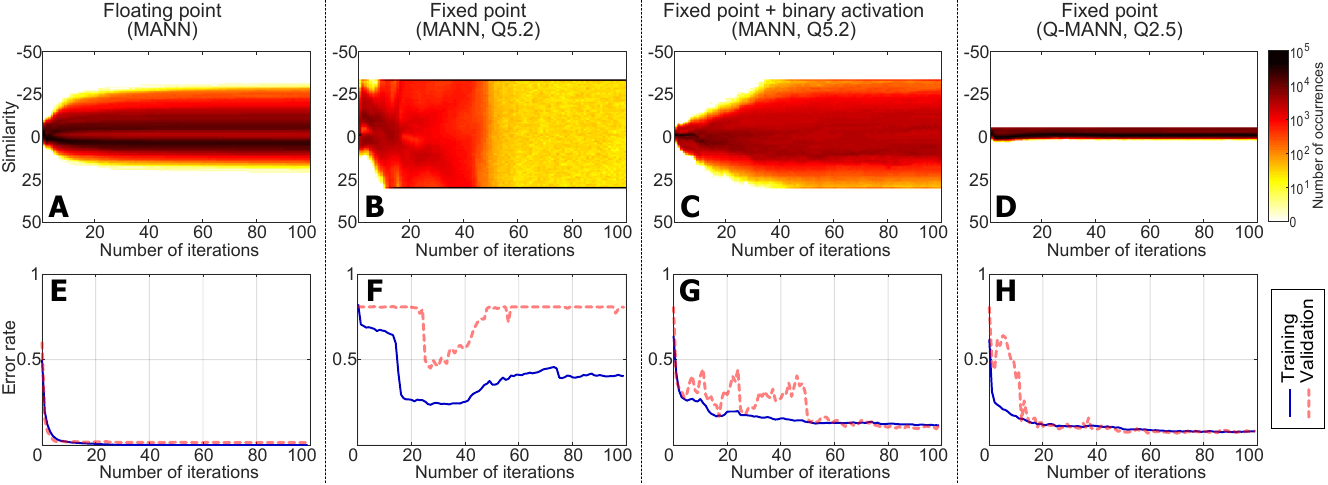}
	\caption{
	The results for MANN and Q-MANN on the bAbI dataset (task8), (A)-(D): the distribution of similarity measure, (E)-(H): the error rate of training and validation. (A) and (E) were the results with 32-bit floating-point. As depicted in (A), the distribution width of cosine similarity became wider as the training progressed.
	In addition, when applying fixed-point quantization, the width became much wider than that of floating point due to the quantization error as (B).
	Thus, this training tendency of cosine similarity caused the fixed-point overflow, which considerably degraded the learning performance when training with fixed-point parameters and activations as (F).
	By applying binary quantization to activations, multiplications can be replaced by additions in cosine similarity, which could prevent numerous fixed-point overflows as (C).
	However, due to the lack of the accuracy in similarity measure, the learning performance was limited as (G).
	Q-MANN using bounded similarity (Hamming similarity) was suitable for applying the fixed-point quantization with limited numerical representation because the distribution width of similarity did not increase as training progressed shown in (D).
	The result of Q-MANN (H) showed a lower error rate than those of MANN (F and G) when applying 8-bit fixed-point quantization.
	}
	\label{fig:total_similarity_lr_curve}
\end{figure*}

In this paper, we proposed Q-MANN to overcome the aforementioned disadvantage of conventional MANN, which is vulnerable to quantization error. As we have seen, fixed-point overflow in content-based addressing with cosine similarity significantly degraded the learning performance of the MANN. Our solution to this problem was using bounded similarity (e.g., Hamming similarity) instead of cosine similarity in content-based addressing. Bounded similarity could prevent fixed-point overflow and improve learning performance when applying quantization. The proposed similarity measure $S_H$ is defined as
\begin{equation}
\label{eq:ham_sim_proposed}
S_H(\hat{U},\hat{V}) 
=\sum _{ i }{ S_{ \hat{u}_{ i } } S_{ \hat{v}_{ i } } \sum _{ k=0 }^{ n-2 }{ W_{ k } XNOR(\hat{u}_{ik},\hat{v}_{ik})}  }\textrm{,}
\end{equation}
where $\hat{U}~\text{and}~\hat{V}$ are the quantized input vectors, $S_{\hat{u}_{i}}~\text{and}~S_{\hat{v}_{i}}$ are the signs, $\hat{u}_{ik}~\text{and}~\hat{v}_{ik}$ are the bits of each element of the input vectors, $W_k$ is the Hamming similarity weight, and $n$ is the bit width of the fixed point.
The proposed vector similarity is measured by adding the Hamming similarity between the elements $\hat{u}_i~\text{and}~\hat{v}_i$ of the fixed-point input vectors $\hat{U}~\text{and}~\hat{V}$ as Equation~\ref{eq:ham_sim_proposed}.

The similarity value between two fixed-point scalars $\hat{u_i}~\text{and}~\hat{v_i}$ should be large as the bits of the integer and fractional parts $\hat{u}_{ik}~\text{and}~\hat{v}_{ik}$ are similar if the signs $S_{\hat{u}_i}~\text{and}~S_{\hat{u}_i}$ of the two scalars are the same, and small if the signs are different. Further, in the case of different signs, the similarity value should be lower than when the signs are the same regardless of the similarity of the other bits. Thus, the signs of the two scalars are multiplied by the Hamming similarity of the bits of the integer and fractional parts as shown in Equation~\ref{eq:ham_sim_proposed}.

Each fixed-point bit represents a different value depending on its position. Hence, for more accurate similarity measurement, the similarity value between two fixed-point scalars was measured with weight. We used the weight as $W_k=2^{k+\alpha-n}$ for the vector similarity measure of Q-MANN without significantly increasing the computation-energy consumption. The weight constant $\alpha$ was empirically determined through various experiments.

For end-to-end training using gradient back-propagation and SGD as a conventional MANN, all components of MANN must be differentiable.
Thus, we smoothed a quantization function for gradient back-propagation while we used it as a step-wise function in forward propagation.
In addition, the Hamming distance is identical to the Manhattan distance for binary vectors and the Manhattan distance is differentiable, the proposed function can be used in end-to-end training by using SGD and gradient back-propagation.
Furthermore, the computation-energy consumption of gradient back-propagation can be lowered through an approximation that uses only addition and shift operations.
Consequently, the proposed gradient back-propagation function is defined as
\begin{equation}
\label{eq:ham_sim_proposed_bwd}
\frac{\partial{S_H(\hat{U},\hat{V})}}{\partial{\hat{u}_i}}
\approx S_{\hat{u}_i} 2^{\alpha} (S_{\hat{u}_i}-S_{\hat{v}_i}) - \sum _{k=0}^{n-2}{ (S_{\hat{v}_i} 2^{\alpha} (\hat{u}_{ik}-\hat{v}_{ik})})\textrm{.}
\end{equation}

As shown in Figure~\ref{fig:QMANN_similarity_lr_curve}A, Q-MANN did not expand the distribution of similarity even when training progressed. To investigate the effect of the quantization error on the distribution of similarity in Q-MANN, $FRAC$ was reduced to 7, 4, and 1 with $IWL$ of 5 and the distribution was measured as in the experiments on the conventional MANN (Figures~\ref{fig:QMANN_similarity_lr_curve}A,~\ref{fig:QMANN_similarity_lr_curve}B, and~\ref{fig:QMANN_similarity_lr_curve}C, respectively).
The results showed that there was no significant difference in the width of the distributions.
They were quite different from those in the experiments on the conventional MANN where the distribution width became wider as $FRAC$ decreased.

To avoid overflow in fixed-point quantization with limited numerical representation range, Q-MANN was optimized for a narrow distribution of vector similarity.
Because of this, even if many bits were used, the accuracy would become poor for a significant gap between the actual width of the distribution and the numerical representation range.
Thus, the use of optimized $IWL$ (Q2.7, Figure~ \ref{fig:QMANN_similarity_lr_curve}H) improved the learning performance compared to using more $IWL$ (Q5.7, Figure~\ref{fig:QMANN_similarity_lr_curve}E).

Since Q-MANN is trained with a narrow distribution of similarity, it can prevent fixed-point overflow. 
This enhances the robustness of the quantization error and stabilizes training by applying fixed-point quantization.
In addition, the similarity measure function of Q-MANN uses only simple operations such as addition, shift, and XNOR gate operations.
Hence, Q-MANN is suitable in a limited-resource environment.

\subsection{Effective training techniques}
One of the other reasons for poor learning performance was in the RNN used as the memory controller when applying fixed-point quantization to MANN.
The RNN uses the same parameters for different time steps. Thus, if the quantization error of the parameter is equal to each time step, the influence of the error can be greatly magnified.
We reduced the influence in the memory controller by employing slightly different $IWL$ and $FRAC$ at every time step while maintaining the bits used for quantization.
We named this method \textit{memory controller quantization control} (MQ).
The use of MQ made it possible to vary the quantization error at every time step without increasing the quantization overhead, such that the error could be canceled each other.

In addition, as shown in Figure~\ref{fig:total_similarity_lr_curve}B, when fixed-point overflow occurred in the similarity measure due to quantization, the training and validation error increased sharply (Figure~\ref{fig:total_similarity_lr_curve}F).
With \textit{early stopping} (ES), we could reduce the degradation and variance of learning performance caused by the quantization error.

\section{Experimental Results and Discussion}

\ctable[
star,
pos = t,
center,
caption = {Test error rates (\%) on the bAbI dataset for various configurations and their computation-energy gain (ES=Early Stopping, MQ=Memory controller Quantization control)},
label = {tab:experimental_result},
doinside = {\footnotesize \def\arraystretch{.7}}
]{lccccccc}{
    \tnote[a]{ calculated by using data from~\cite{horowitz20141}}
}{
    \toprule
	& & \multicolumn{3}{c}{\multirow{2}{*}{bit-width}} & \multicolumn{2}{c}{\multirow{2}{*}{ error rate (\%)}} & computation \\
	
	& & \multicolumn{3}{c}{} & \multicolumn{2}{c}{} & {-energy} \\
	
	\midrule
	\multirow{2}{*}{} & \multirow{2}{*}{type} & \multirow{2}{*}{input} & \multirow{2}{*}{parameter} & \multirow{2}{*}{activation} & avg. & avg. & \multirow{2}{*}{gain\tmark[a]} \\
	& & & & & of best & of mean &  \\
	\midrule
	\midrule
	MANN & floating & 1 & 32 & 32 & 15.33 & 17.14 & 1$\times$ \\
	
	\midrule
	
	MANN & fixed & 1 & 8 & 8 & 40.04 & 51.23 & 20.22$\times$ \\
	MANN + ES & fixed & 1 & 8 & 8 & 39.31 & 43.87 & 20.22$\times$ \\
	MANN + ES + MQ & fixed & 1 & 8 & 8 & 33.67 & 38.50 & 20.22$\times$ \\
	
	Q-MANN & fixed & 1 & 8 & 8 & 24.33 & 30.43 & 21.03$\times$ \\
	Q-MANN + ES & fixed & 1 & 8 & 8 & 26.47 & 32.88 & 21.03$\times$ \\
	Q-MANN + ES + MQ & fixed & 1 & 8 & 8 & \textbf{22.30} & \textbf{27.68} & 21.03$\times$ \\
	
	\midrule
	
	MANN & fixed & 1 & 8 & 1 & 30.81 & 44.49 & 22.25$\times$ \\
	MANN + ES & fixed & 1 & 8 & 1 & 31.41 & 37.12 & 22.25$\times$\\
	MANN + ES + MQ & fixed & 1 & 8 & 1 & 30.14 & 42.24 & 22.25$\times$ \\
		
	Q-MANN & fixed & 1 & 8 & 1 & 27.11 & 31.91 & 22.25$\times$ \\
	Q-MANN + ES & fixed & 1 & 8 & 1 & 26.66 & 32.66 & 22.25$\times$ \\
	Q-MANN + ES + MQ & fixed & 1 & 8 & 1 & \textbf{25.63} & \textbf{31.00} & 22.25$\times$ \\
	\bottomrule   
}

We verified the performance of the proposed Q-MANN and training method by applying quantization.
All experiments were performed on the bAbI dataset (10k)~\cite{weston2015towards}, by using a training method and hyperparameters similar to those in~\cite{sukhbaatar2015end}.
Details of the model and hyperparameters are provided in Supplementary Table~\ref{tab:sup_model_descriptions} and Table~\ref{tab:sup_hyperparameters}.

In a manner similar to that in~\cite{sukhbaatar2015end}, we repeated each training 10 times to obtain the best and mean error rate of each task in the bAbI dataset. 
We used the average of the best and mean error rates for each task as the two types of performance measuring metrics.

Optimal fixed-point quantization for $IWL$ and $FRAC$ were obtained through experiments. The optimal learning performance of the 8-bit fixed-point quantization was obtained when using Q5.2 for conventional MANN and Q2.5 for Q-MANN.

Table~\ref{tab:experimental_result} provides the experimental results under various conditions.
The entire results are included in Supplementary Table~\ref{tab:sup_result_fixed} and Table~\ref{tab:sup_result_fixed_bin_act}.
The binary vector was used as the input in the form of Bag-Of-Words for the computation efficiency. The baseline configuration of our experiments adopted and 32-bit floating-point parameters and activations with the binary input.

We obtained a gain of about 20$\times$ in computational energy compared to the baseline when we applied 8-bit fixed-point quantization (Q5.2) to the conventional MANN. However, the average error rates of the best and mean cases increased by 160\% and 200\%, respectively. The learning performance significantly deteriorated (Figure~\ref{fig:total_similarity_lr_curve}B)  because the quantization error greatly increased due to the fixed-point overflow in the similarity measure (Figure~\ref{fig:total_similarity_lr_curve}F). We achieved an increase of 159\% in the average error rate of the mean case by applying ES. The error rate was increased by 120\% by applying both ES and MQ compared to the baseline. 

In the case of Q-MANN, 8-bit fixed-point quantization (Q2.5) resulted in a gain of about 21$\times$ in computational energy compared to the baseline and the average error rate of the best and mean cases were reduced by about 37\% and 41\%, respectively, compared with that of the conventional MANN. 
As shown in Figure~\ref{fig:total_similarity_lr_curve}D, these results can be attributed to the infrequent occurrence of fixed-point overflow at a similarity measure in Q-MANN, hence it became robust to the quantization error trained as in Figure~\ref{fig:total_similarity_lr_curve}H.
As a result of applying ES and MQ to Q-MANN, the error rate was reduced by up to 46\% compared with the case of conventional MANN.

We applied binary quantization to activations of the conventional MANN with 8-bit fixed-point quantized parameters.
As a result, we obtained a gain in computational energy of about 22$\times$ compared to that of the baseline.
In addition, the average error rate of the best and mean cases increased by about 100\% and 160\% compared to the baseline, which showed a lower increase in the error rate compared to that of 8-bit fixed-point quantization. 
This was because the overflows were reduced by binarization as in Figure~\ref{fig:total_similarity_lr_curve}C.
Hence, the quantization error caused by fixed-point overflow had a greater impact on the learning performance of conventional MANN than the lack of information due to binarization.
As a result of training with binarization, the error rate of Q-MANN with ES and MQ could be reduced by up to 17\% and 30\%, respectively, compared to the conventional MANN because Q-MANN could use enough information while preventing fixed-point overflow.

The ultimate goal of this study is to train a MANN in a limited-resource environment through quantization, which allows us to use dot product instead of cosine similarity as an approximation to reduce the amount of computational energy consumed.
Our analysis showed that conventional MANN with dot product is vulnerable to a quantization error when trained.
Since cosine similarity is the normalized value of the dot product, it implies that a MANN with cosine similarity is also vulnerable to a quantization error.
Hence, the advantage of Q-MANN proposed in this paper is still valid using cosine similarity.

\section{Conclusion}
In this paper, we applied fixed-point and binary quantization to conventional MANN in both of training and inference as a pioneering study.
Through theoretical analysis and various experiments, we revealed that the quantization error had a great impact on the learning performance of conventional MANN.
Based on our in-depth analysis, we proposed Q-MANN which use bounded similarity in content-based addressing, which is suitable for the fixed-point and binary quantization with a limited numerical representation.
We also showed that Q-MANN could be converged and achieve more robust learning performance in comparison to a conventional MANN for fixed-point and binary quantization using a few bits.


\bibliographystyle{aaai}
\bibliography{aaai_2018}

\newpage
\onecolumn
\section*{Supplementary Information}
\renewcommand\thesection{\Alph{section}}
\setcounter{section}{0}

\section{Fixed-Point Quantization and Computational Energy}
\subsection{Fixed-point quantization} \label{sup_fixed_point}

\begin{equation*}
    \begin{split}
        U,V&:\textrm{ floating-point vector}\\ 
        \hat{U},\hat{V}&:\textrm{ fixed-point vector}\\
        \hat{u}_i,\hat{v}_i&:\textrm{ fixed-point scalar (binary vector)} \in{\left\{0,1 \right\}}^{n} \\
        \hat{u}_{ik},\hat{v}_{ik}&\in \left\{0,1 \right\} \\
        \epsilon_{\hat{u}_i}&:\textrm{ quantization error of }\hat{u}_i \\
    \end{split}
\end{equation*}

\begin{equation}
    \begin{split}
        \hat{u}_i &= u_i+\epsilon_{\hat{u}_i} \\
        &= sign 2^{-FRAC} \sum _{k=0}^{n-2}{2^k \hat{u}_{ik}} \\
    \end{split}
\end{equation}

\begin{equation}
\label{eq:sup_fixed_point_quant_error}
    |\epsilon_{u_i}| < 
    \begin{cases}
        2^{-FRAC}         & \textrm{if}~|u_i|<2^{IWL} \\ 
        |2^{IWL}-|u_i||   & \textrm{if}~|u_i|\geq2^{IWL} \textrm{ (fixed-point overflow occurs)}
    \end{cases}
\end{equation}

\subsection{Computational energy}
\ctable[
pos = h,
caption = {Computation-energy gain from~\cite{horowitz20141}},
label = {tab:sup_arithmetic_operation_energy},
]{lcccc}{
\tnote[a]{ compared with 32-bit floating-point mult }
}{ \FL
    Type & Arithmetic operation & Bit & Energy (pJ) & Gain\tmark[a] \ML
    Fixed point & add & 8 & 0.03 & 123.3 \NN
	& & 32 & 0.1 & 37 \NN
	& mult & 8 & 0.2 & 18.5 \NN
	& & 32 & 3.1 & 1.2 \ML
	
	Floating point & add & 16 & 0.4 & 9.3\NN
	& & 32 & 0.9 & 4.1 \NN
	& mult & 16 & 1.1 & 3.4 \NN
	& & 32 & 3.7 & 1 \LL
}

\section{Experimental Results} \label{sup_exp_results}

\ctable[
sideways,
pos=ph,
caption = {
Test error rates (\%) of repeating each training 10 times for 8-bit fixed-point quantization on the bAbI dataset (COS=COSine similarity, HAM=HAMming similarity, ES=Early Stopping, MQ=Memory controller Quantization control)},
label = {tab:sup_result_fixed},
doinside = {\scriptsize \def\tabcolsep{3.2pt} \def\arraystretch{1.4}}
]{rlrrrrrrrrrrrrrrrrrrrrr}{}
{\FL
&& \multicolumn{12}{c}{MANN} & \multicolumn{9}{c}{Q-MANN} \NN
\cmidrule(lr){3-14}\cmidrule(lr){15-23}
\multicolumn{2}{l}{numerical representation}&\multicolumn{3}{c}{floating point}&\multicolumn{3}{c}{fixed point (Q5.2)}&\multicolumn{3}{c}{fixed point (Q5.2)}&\multicolumn{3}{c}{fixed point (Q5.2)}&\multicolumn{3}{c}{fixed point (Q2.5)}&\multicolumn{3}{c}{fixed point (Q2.5)}&\multicolumn{3}{c}{fixed point (Q2.5)}\NN
\multicolumn{2}{l}{similarity measure function}&\multicolumn{3}{c}{COS}&\multicolumn{3}{c}{COS}&\multicolumn{3}{c}{COS}&\multicolumn{3}{c}{COS}&\multicolumn{3}{c}{HAM}&\multicolumn{3}{c}{HAM}&\multicolumn{3}{c}{HAM}\NN
&&&&&&&&\multicolumn{3}{c}{ES}&\multicolumn{3}{c}{ES+MQ}&&&&\multicolumn{3}{c}{ES}&\multicolumn{3}{c}{ES+MQ}\NN[1.0ex]
\cmidrule(lr){1-2}\cmidrule(lr){3-5}\cmidrule(lr){6-8}\cmidrule(lr){9-11}\cmidrule(lr){12-14}\cmidrule(lr){15-17}\cmidrule(lr){18-20}\cmidrule(lr){21-23}
Task&&min&mean&std&min&mean&std&min&mean&std&min&mean&std&min&mean&std&min&mean&std&min&mean&std\NN
\cmidrule(lr){1-2}\cmidrule(lr){3-5}\cmidrule(lr){6-8}\cmidrule(lr){9-11}\cmidrule(lr){12-14}\cmidrule(lr){15-17}\cmidrule(lr){18-20}\cmidrule(lr){21-23}
1&1 supporting fact&0.0&0.00&0.000&0.0&16.25&34.259&0.0&2.09&5.274&0.0&1.19&0.936&0.1&0.97&0.801&0.3&0.73&0.483&0.4&1.01&0.465\NN
2&2 supporting facts&0.2&0.55&0.217&79.6&81.46&1.494&75.1&80&2.395&44.8&66.65&7.917&20.3&51.33&22.004&22.8&58.49&22.471&20.9&42.94&17.837\NN
3&3 supporting facts&23.1&25.47&1.282&78.5&80.71&1.942&78.5&82.11&3.198&72.3&76.58&2.686&75.7&79.77&2.739&74.8&77.5&1.671&53.7&68.28&5.923\NN
4&2 argument relations&30.8&32.08&1.049&31.6&33.03&1.370&31.3&32.67&0.730&40.7&48.83&5.997&31.4&32.77&1.656&30.5&33.23&3.774&30.6&35.8&6.806\NN
5&3 argument relations&13.5&15.08&0.750&69.6&76.6&6.130&68&70.14&0.969&39.8&46.82&7.262&13&14.66&1.152&12.6&15.26&1.581&15.3&16.44&0.635\NN
6&yes/no questions&3.1&5.23&2.591&49.3&49.94&0.389&47.9&49.57&0.617&47.7&49.58&0.807&15.1&17.81&2.433&15.6&20.64&4.160&12.1&15.56&2.334\NN
7&counting&11.4&12.63&1.041&51.1&52.83&0.790&22.1&30.97&8.749&21.7&23.04&1.227&21.3&24.57&3.192&20.8&21.8&0.872&17.8&19.3&1.166\NN
8&lists/sets&0.6&1.34&0.406&36.7&73.92&13.713&53.7&59.18&6.187&12&22.95&9.002&7.3&10.9&2.390&6.7&9.75&1.967&9.2&10.56&0.956\NN
9&simple negation&3.9&6.42&2.280&35.1&36.47&0.796&36.1&36.28&0.123&35.8&36.16&0.126&29.7&35.21&2.096&28.5&35.27&2.468&14.4&16.83&2.502\NN
10&indefinite knowledge&7.2&10.42&2.206&57.4&66.12&9.353&55.9&57.07&1.214&52.9&56.04&1.617&36.4&48.55&6.415&45.8&52.04&4.006&26.4&28.9&1.648\NN
11&basic coreference&0.2&9.11&5.572&11&31.82&25.406&9.6&10.85&0.712&9&11.81&1.735&1.9&10.72&3.485&1.7&13.66&5.571&7.5&11.69&2.788\NN
12&conjunction&0.0&0.00&0.000&0.0&16.82&35.435&0.0&0.43&1.290&0.0&8.25&10.995&0.0&0.29&0.260&0.0&5.7&17.007&0.0&0.6&0.467\NN
13&compound coreference&0.0&0.16&0.207&8.7&59.99&24.820&5.6&37.19&31.030&5.6&5.74&0.165&5.7&7.86&2.450&19.3&33.26&11.866&0.0&5.87&2.882\NN
14&time reasoning&3.3&3.90&0.313&5.5&20.18&21.807&4.7&20&16.633&9.8&17.78&3.331&14.1&15.56&1.083&13.1&15.13&1.084&13.9&28.33&13.601\NN
15&basic deduction&10.7&13.48&1.850&51.4&54.78&2.193&49.9&53.21&2.275&51&54.51&2.315&14.8&28.66&15.171&15.4&37.69&12.936&15.8&33.52&10.069\NN
16&basic induction&50.8&53.19&1.537&51.5&58.27&4.649&53&56&3.497&49.7&53.01&2.175&50.7&52.6&1.341&50.7&51.77&0.657&49.6&51.95&1.372\NN
17&positional reasoning&45&46.56&1.246&47.8&48.78&1.698&52&52&0.000&48&51.2&1.687&37.6&39.71&2.591&37.7&38.95&0.826&37.5&38.88&1.107\NN
18&size reasoning&38.4&42.32&2.185&46.2&48.58&2.553&52.9&53.57&0.564&42.4&47.43&3.952&41.5&45.85&2.570&42.9&45.96&1.893&40&43.21&1.596\NN
19&path finding&64.3&64.86&0.350&89.8&90.82&0.924&89.8&90.63&0.523&90.1&91.24&0.554&90&90.89&0.702&90.2&90.78&0.379&80.8&83.93&2.074\NN
20&agent’s motivation&0.0&0.00&0.000&0.0&27.26&36.234&0.0&3.4&3.723&0.0&1.16&1.083&0.0&0.0&0.000&0.0&0.0&0.000&0.0&0.0&0.000\ML
&Average error (\%)&15.325&17.14&1.254&40.040&51.232&11.298&39.305&43.868&4.485&33.665&38.499&3.278&25.330&30.434&3.727&26.470&32.881&4.784&22.295&27.680&3.811\LL
}

\ctable[
sideways,
pos=ph,
caption = {
Test error rates (\%) of repeating each training 10 times for 8-bit fixed-point and binary quantization on the bAbI dataset (COS=COSine similarity, HAM=HAMming similarity, ES=Early Stopping, MQ=Memory controller Quantization control)
},
label = {tab:sup_result_fixed_bin_act},
doinside = {\scriptsize \def\tabcolsep{3.2pt} \def\arraystretch{1.4}}
]{rlrrrrrrrrrrrrrrrrrrrrr}{}
{\FL
&& \multicolumn{12}{c}{MANN} & \multicolumn{9}{c}{Q-MANN} \NN
\cmidrule(lr){3-14}\cmidrule(lr){15-23}
\multicolumn{2}{l}{numerical representation}&\multicolumn{3}{c}{floating point}&\multicolumn{3}{c}{fixed point (Q5.2)}&\multicolumn{3}{c}{fixed point (Q5.2)}&\multicolumn{3}{c}{fixed point (Q5.2)}&\multicolumn{3}{c}{fixed point (Q2.5)}&\multicolumn{3}{c}{fixed point (Q2.5)}&\multicolumn{3}{c}{fixed point (Q2.5)}\NN
\multicolumn{2}{l}{similarity measure function}&\multicolumn{3}{c}{COS}&\multicolumn{3}{c}{COS}&\multicolumn{3}{c}{COS}&\multicolumn{3}{c}{COS}&\multicolumn{3}{c}{HAM}&\multicolumn{3}{c}{HAM}&\multicolumn{3}{c}{HAM}\NN
&&&&&&&&\multicolumn{3}{c}{ES}&\multicolumn{3}{c}{ES+MQ}&&&&\multicolumn{3}{c}{ES}&\multicolumn{3}{c}{ES+MQ}\NN[1.0ex]
\cmidrule(lr){1-2}\cmidrule(lr){3-5}\cmidrule(lr){6-8}\cmidrule(lr){9-11}\cmidrule(lr){12-14}\cmidrule(lr){15-17}\cmidrule(lr){18-20}\cmidrule(lr){21-23}
Task&&min&mean&std&min&mean&std&min&mean&std&min&mean&std&min&mean&std&min&mean&std&min&mean&std\NN
\cmidrule(lr){1-2}\cmidrule(lr){3-5}\cmidrule(lr){6-8}\cmidrule(lr){9-11}\cmidrule(lr){12-14}\cmidrule(lr){15-17}\cmidrule(lr){18-20}\cmidrule(lr){21-23}
1&1 supporting fact&0.0&0.00&0.000&2.5&58.27&25.378&0.0&0.18&0.193&0.0&2.9&1.058&0.3&2.42&4.143&1.1&1.34&0.178&1.3&2.7&1.738\NN
2&2 supporting facts&0.2&0.55&0.217&61.5&69.4&4.046&65.8&72.21&5.024&19.3&74.7&17.069&34.5&55.92&12.417&38.6&60.93&11.992&27.9&49.54&16.441\NN
3&3 supporting facts&23.1&25.47&1.282&79&81.68&2.125&78.5&83.12&2.603&73.3&77.9&1.335&68.8&71.1&1.409&68.7&70.57&1.008&65.5&71.23&2.318\NN
4&2 argument relations&30.8&32.08&1.049&34.6&45.55&10.995&35.6&39.86&2.449&35.3&61.2&7.573&42.2&47.37&3.144&44.5&49.16&3.356&41&44.84&2.056\NN
5&3 argument relations&13.5&15.08&0.750&32.6&60.05&15.753&28.4&43.9&13.898&34.5&57.6&8.849&19.1&23.19&4.021&16.6&19.27&2.400&17.1&19.04&1.065\NN
6&yes/no questions&3.1&5.23&2.591&22.9&40.45&10.501&24.5&37.89&10.597&48.1&52&1.079&14.9&18.7&4.225&13.4&19.58&8.248&14.2&22.8&6.186\NN
7&counting&11.4&12.63&1.041&22.3&29.95&5.713&22&24.54&2.448&18&26.5&2.507&17.6&19.07&1.253&17.7&19.23&1.352&18.3&20.01&1.031\NN
8&lists/sets&0.6&1.34&0.406&20.8&37.61&12.054&16.4&34.42&8.679&9.4&33.7&8.850&10.7&14.79&7.396&10.3&18.07&9.433&10.6&11.62&0.939\NN
9&simple negation&3.9&6.42&2.280&21.6&28.16&4.146&26.3&29.3&3.765&23.3&28.6&1.611&13.2&16.22&1.550&12.2&17.33&2.261&14.9&16.89&1.610\NN
10&indefinite knowledge&7.2&10.42&2.206&41.9&47.04&4.743&44.8&46.29&2.355&42.1&50.7&2.149&28.2&37.8&5.173&31.4&37.05&3.623&27.4&32&2.978\NN
11&basic coreference&0.2&9.11&5.572&11.1&26.33&22.571&10.3&11.85&1.417&9.3&34.5&7.502&9.6&12.11&1.189&11.1&13.64&2.926&10.9&12.23&1.163\NN
12&conjunction&0.0&0.00&0.000&0.0&6.93&14.217&0.0&0.06&0.084&0.0&3.2&1.043&1.1&3.49&1.994&1.4&16.28&15.443&1.2&7.7&10.675\NN
13&compound coreference&0.0&0.16&0.207&5.6&47.08&29.219&5.6&5.6&0.000&5.4&8.6&1.165&4.9&6.45&1.439&4&6.55&1.912&0.2&5.51&2.204\NN
14&time reasoning&3.3&3.90&0.313&10.2&14.85&4.348&8&14.93&6.156&10.6&19&2.615&31.5&40.02&4.653&28.3&39.04&7.043&27.2&35.6&7.773\NN
15&basic deduction&10.7&13.48&1.850&23&52.78&10.862&30.5&54.18&8.635&47.6&57.3&2.629&29.8&44.67&5.847&19.8&39.56&11.884&17.5&42.48&9.933\NN
16&basic induction&50.8&53.19&1.537&53&56.29&2.725&54.8&55.91&1.093&51.4&58.4&2.299&50.7&52.23&1.480&50.7&52.17&1.081&49.7&52.06&1.265\NN
17&positional reasoning&45&46.56&1.246&44.5&47.33&2.110&47.8&49.33&1.140&49&52.6&1.036&37.4&39.64&1.796&36.6&40.09&2.938&38.6&40.62&1.971\NN
18&size reasoning&38.4&42.32&2.185&44.7&47.24&2.689&44.8&50.01&3.609&42.8&51.1&2.482&41.3&45.19&1.846&42.7&45.95&1.317&44.1&45.89&1.658\NN
19&path finding&64.3&64.86&0.350&84.4&87.42&2.410&84.1&86.94&2.482&83.4&90.1&1.810&86.4&87.54&0.929&84.1&87.25&1.325&84.9&87.15&1.023\NN
20&agent’s motivation&0.0&0.00&0.000&0.0&5.34&6.901&0.0&1.81&1.783&0.0&4.1&1.335&0.0&0.18&0.210&0.0&0.04&0.126&0.0&0.03&0.067\ML
&Average error (\%)&15.325&17.14&1.254&30.810&44.488&9.675&31.410&37.117&3.921&30.140&42.235&3.800&27.110&31.905&3.306&26.660&32.655&4.492&25.625&30.997&3.705\LL
}

\newpage
\section{MANN Model Description} \label{sup_mann_model}
\ctable[
pos=h,
caption = {Model Descriptions},
label = {tab:sup_model_descriptions},
]{cll}{
}{  \FL
	Symbol & Description & Domain \ML
	\(I\)   & dimension of input & \(\mathbb{N}\)     \NN
	\(E\)   & dimension of internal representation & \(\mathbb{N}\)     \NN
	\(L\)   & number of memory element & \(\mathbb{N}\)     \NN
	\(R\)   & number of read & \(\mathbb{N}\)     \NN
	
	\(V\)   & input vectors (sentences) & \(\mathbb{R}^{I\times L}\)     \NN
	\(q\)   & input vector (question) & \(\mathbb{R}^{I}\)     \NN
	
	\(W_a\) & weight of input(V) & \(\mathbb{R}^{E\times I}\)     \NN
	\(W_q\) & weight of input(q) & \(\mathbb{R}^{E\times I}\)     \NN
	\(W_r\) & weight of read memory & \(\mathbb{R}^{E\times I}\)     \NN
	\(W_k\) & weight of read key & \(\mathbb{R}^{E\times E}\)     \NN
	\(W_o\) & weight of output & \(\mathbb{R}^{I\times E}\)     \NN
	
	\(M_a\) & address memory & \(\mathbb{R}^{E\times L}\)     \NN
	\(M_r\) & read memory & \(\mathbb{R}^{E\times L}\)     \NN
	
	\(k_i\) & \(i\)th read key (\(1 \leq i \leq R\)) & \(\mathbb{R}^E\)   \NN
	\(w_{r,i}\) & \(i\)th read weight & \(\mathbb{R}^{L}\)     \NN
	
	\(r_{i}\) & \(i\)th read vector & \(\mathbb{R}^{E}\)     \NN
	\(o_{i}\) & \(i\)th output vector & \(\mathbb{R}^{I}\)     \LL
}

\begin{equation*}
    \label{eq:sup_model}
    \begin{split}
        \text{Memory addressing (content-based): } & \\
        S(u,v) &= {u\cdot v} \\
        C(M,k)[i] &= \frac{exp\{S(M_i,k)\}}{\sum_{j}^{L}{exp\{S(M_j,k)\}}} \\
        \text{Memory update: } & \\
        M_a &= W_a V \\
        M_r &= W_r V \\
        \text{Memory read: } & \\
        k^i &=
        \begin{cases}
            W_q q & \textrm{if}~i=1 \\
            W_k k_{i-1} + r_i & \textrm{otherwise} \\
        \end{cases} \\   
        w_{r,i} &= C(M_a,k_i) \\
        r_i &= M_r w_{r,i} \\
        \text{Output: } & \\
        o_i &= softmax(W_o k_i) \\
    \end{split}
\end{equation*}

\section{Analysis of the Effect of Quantization Error on Conventional MANN}
\subsection{Vector similarity measure function - dot product}\label{dot_product}


\begin{equation}
    \label{eq:sup_quant_effect_dot_product}
    \begin{split}
        \hat{Z} &=\hat{U}\cdot \hat{V} \\
        &=\sum _{}^{}{\hat{u}_i \hat{v}_i} \\
        &=\sum {(u_i+\epsilon_{u_i})(v_i+\epsilon_{v_i})} \\
        &=\sum {u_i v_i} + \sum {(u_i \epsilon_{v_i} + v_i \epsilon_{u_i})} + \sum {\epsilon_{u_i} \epsilon_{v_i}} \\
        &\approx \sum {u_i v_i} + \sum {(u_i \epsilon_{v_i} + v_i \epsilon_{u_i})} \\
        &=Z + \epsilon_{Z}
    \end{split}
\end{equation}

\subsection{Normalization function - Softmax}\label{softmax}

\begin{equation}
    \label{eq:sup_quant_effect_softmax}
    \begin{split}
        \hat{y}_i &= \frac{\exp(\hat{z}_i)}{\sum {\exp(\hat{z}_k)}} \\
        &=\frac{\exp(z_i+\epsilon_{z_i})}{\sum {\exp(z_k+\epsilon_{z_k})}} \\
        &=\frac{\exp(z_i)}{\sum {\exp(z_k+\epsilon_{z_k}-\epsilon_{z_i})}} \\
        &\leq\frac{\exp(z_i)}{\sum {\exp(z_k-\epsilon_{max})}} \\
        &=\frac{\exp(z_i)}{\exp(-\epsilon_{max}) \sum {\exp(z_k)}} \\
        &=\exp(\epsilon_{max})y_i 
    \end{split}
\end{equation}

\section{Hyperparameters} \label{sup_hyperpara}
\ctable[
pos=h,
caption = {Hyperparameters},
label = {tab:sup_hyperparameters},
]{lc}{
\tnote[a]{depend on the task in the bAbI dataset}
}{  \FL
	Parameter & Value \ML
	dimension of input ($I$) & 17 - 98\tmark[a] \NN
	dimension of internal representation ($E$) & 60 \NN
	number of memory locations ($L$) & 50 \NN
	number of read ($R$) & 3 \NN
	learning rate & 0.3 \NN
    Hamming similarity weight constant ($\alpha$) & $-3$ \LL
}

\end{document}